\documentclass[letterpaper]{article} % DO NOT CHANGE THIS
\usepackage{aaai2026}  % DO NOT CHANGE THIS
\usepackage{times}  % DO NOT CHANGE THIS
\usepackage{helvet}  % DO NOT CHANGE THIS
\usepackage{courier}  % DO NOT CHANGE THIS
\usepackage[hyphens]{url}  % DO NOT CHANGE THIS
\usepackage{graphicx} % DO NOT CHANGE THIS
\usepackage{natbib}  % DO NOT CHANGE THIS AND DO NOT ADD ANY OPTIONS TO IT
\usepackage{caption} % DO NOT CHANGE THIS AND DO NOT ADD ANY OPTIONS TO IT
\usepackage{algorithm}
\usepackage{algorithmic}

\usepackage{amsfonts}
\usepackage{amsmath}
\usepackage{booktabs}

\usepackage{xcolor}

\usepackage{newfloat}
\usepackage{listings}
\DeclareCaptionStyle{ruled}{labelfont=normalfont,labelsep=colon,strut=off} % DO NOT CHANGE THIS
\floatstyle{ruled}
\newfloat{listing}{tb}{lst}{}
\floatname{listing}{Listing}
\title{MR-COSMO: Visual-Text Memory Recall and Direct CrOSs-MOdal Alignment Method for Query-Driven 3D Segmentation}
\author{
    Chade Li\textsuperscript{\rm 1,2},
    Pengju Zhang\textsuperscript{\rm 1}\footnote{Corresponding authors},
    Yihong Wu\textsuperscript{\rm 1,2}\footnotemark[1]%\thanks{Corresponding authors}
}
\affiliations{
    \textsuperscript{\rm 1}State Key Laboratory of Multimodal Artificial Intelligence Systems,\\Institute of Automation, Chinese Academy of Sciences, Beijing, China.\\
    \textsuperscript{\rm 2}School of Artificial Intelligence, University of Chinese Academy of Sciences, Beijing, China.
    lichade2021@ia.ac.cn, \{pengju.zhang@ia, yhwu@nlpr.ia\}.ac.cn
}

\usepackage{bibentry}
\begin{document}

\maketitle

\begin{abstract}
The rapid advancement of vision-language models (VLMs) in 3D domains has accelerated research in text-query-guided point cloud processing, though existing methods underperform in point-level segmentation due to inadequate 3D-text alignment that limits local feature-text context linking. To address this limitation, we propose \textbf{MR-COSMO}, a Visual-Text \textbf{M}emory \textbf{R}ecall and Direct \textbf{C}r\textbf{OS}s-\textbf{MO}dal Alignment Method for Query-Driven 3D Segmentation, establishing explicit alignment between 3D point clouds and text/2D image data through a dedicated direct cross-modal alignment module while implementing a visual-text memory module with specialized feature banks. This direct alignment mechanism enables precise fusion of geometric and semantic features, while the memory module employs specialized banks storing text features, visual features, and their correspondence mappings to dynamically enhance scene-specific representations via attention-based knowledge recall. Comprehensive experiments across 3D instruction, reference, and semantic segmentation benchmarks confirm state-of-the-art performance.
% The rapid advancement of vision-language models (VLMs) on 3D has driven significant interest in text query-guided point cloud processing tasks. However, existing methods often underperform in point-level tasks like segmentation due to the lack of direct 3D-text alignment, limiting their ability to link local 3D features with textual context. To bridge this gap, we propose \textbf{MR-COSMO}, a Visual-Text \textbf{M}emory \textbf{R}ecall and Direct \textbf{C}r\textbf{OS}s-\textbf{MO}dal Alignment Method for Query-Driven 3D Segmentation, introducing a direct cross-modal alignment module to achieve explicit alignment between 3D point clouds and textual/2D image data. Within the memory module, multiple dedicated banks separately store text features, visual features, and their cross-modal correspondence mappings. These banks are dynamically leveraged through self-attention and cross-attention to update scene-specific features using prior stored knowledge, effectively addressing inconsistencies in query-driven segmentation results across diverse scenarios. Experiments on multiple 3D instruction, reference, and semantic segmentation datasets demonstrate that the proposed method achieves state-of-the-art performance.
\end{abstract}

%     % \link{Code}{https://aaai.org/example/code}
%     % \link{Datasets}{https://aaai.org/example/datasets}
%     \link{Extended version}{https://arxiv.org/abs/2506.20991}

\section{Introduction}
% \label{sec:intro}
The emergence of large language models (LLMs) and 2D visual foundation models (VFMs) has propelled text-guided 3D segmentation to the forefront due to its significant practical implications. This technology aims to segment 3D objects or scenes using natural language inputs. Existing methods \cite{pointclip,seal,gpt4point,segpoint,3dvla,partdistill} typically employ LLMs to interpret textual inputs and leverage the CLIP model \cite{clip} to establish cross-modal associations between 2D images and text. Alternatively, these approaches utilize VFMs to process 2D visual data, relying on camera intrinsic and extrinsic parameters to establish geometric correspondences between 2D projections and 3D coordinates, thereby indirectly inferring 3D point cloud semantics through multi-view fusion.
% The emergence of large language models (LLMs) and 2D visual foundation models (VFMs) has propelled text-guided 3D segmentation to prominence due to its practical significance, aiming to segment 3D objects or scenes using natural language inputs. Existing methods \cite{pointclip,seal,gpt4point,segpoint,3dvla,partdistill} typically employ LLMs to interpret textual inputs and leverage the CLIP model \cite{clip} to establish cross-modal associations between 2D images and text. Alternatively, they utilize VFMs to process 2D visual data, relying on camera intrinsic and extrinsic parameters to establish geometric correspondences between 2D projections and 3D coordinates, indirectly inferring 3D point cloud semantics through multi-view fusion.

\begin{figure}[t]
\centering
\includegraphics[width=0.95\linewidth]{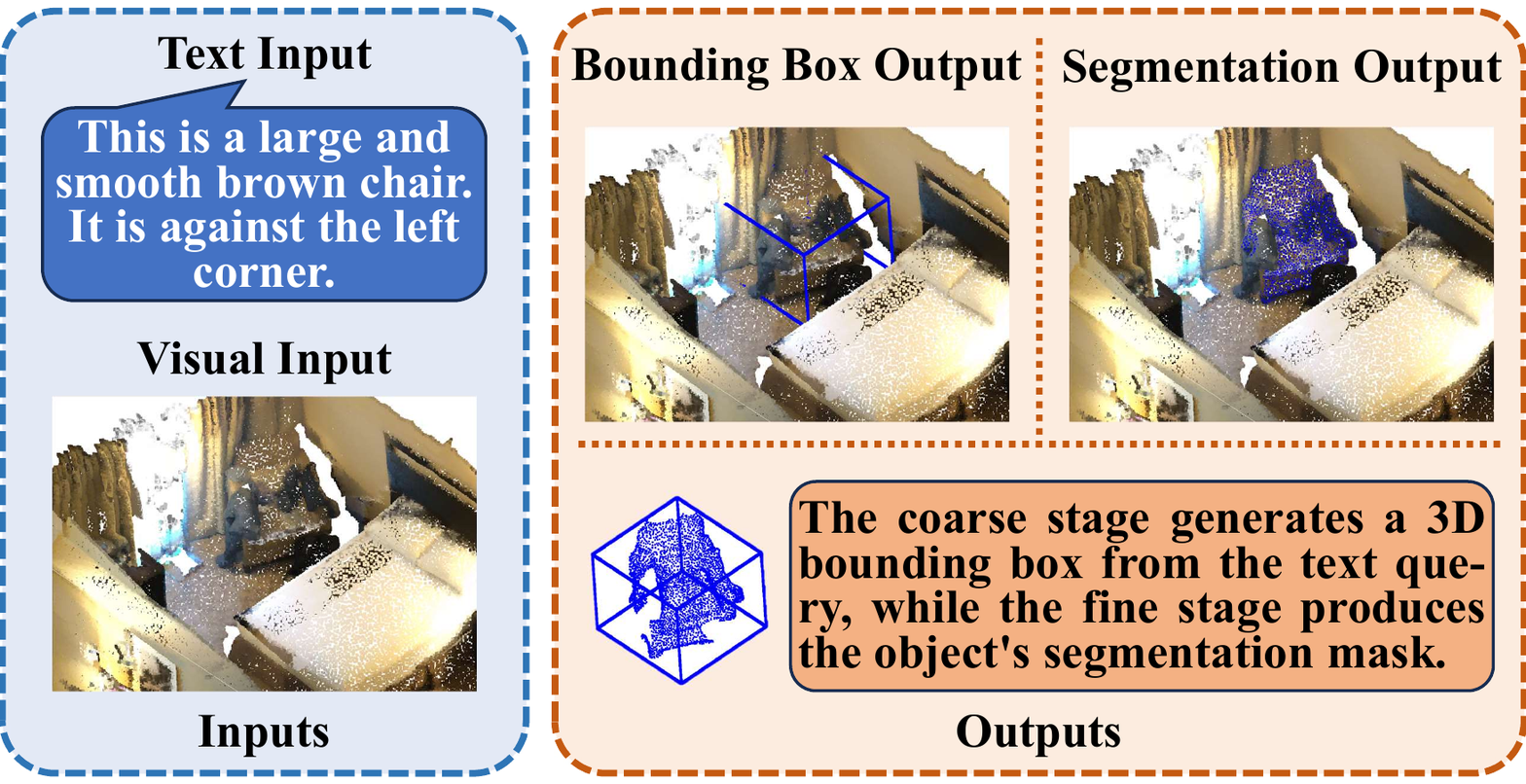}
\caption{The inputs and outputs of the proposed coarse-to-fine query-driven 3D segmentation model.}
\label{fig1}
\end{figure}

PointCLIP \cite{pointclip} extends the understanding of 3D data by utilizing CLIP's \cite{clip} framework to construct alignment between 2D images and text. The CLIP \cite{clip} trains an image encoder and a text encoder through contrastive learning, ensuring that matching image-text pairs converge in the embedding space while non-matching pairs diverge. Seal \cite{seal} employs VFMs for automotive point cloud segmentation, distilling semantic perceptions from VFMs to point clouds via a hyperpixel-driven contrastive learning approach on camera views. However, fine-grained segmentation requires recognizing subtle structural variations within objects, demanding a profound understanding of 3D geometries alongside the capability to capture local details and textual context. The aforementioned methods \cite{pointclip,seal} rely on indirect alignment strategies that use 2D images as intermediaries between 3D point clouds and other modalities. This approach is highly susceptible to errors in the computation of intrinsic/extrinsic parameters and to pixel-point misalignment artifacts. Consequently, existing methods fail to establish stable, accurate coordinate correspondence between 3D point clouds and 2D images, thus lacking the capability required for point-level fine-grained tasks.
% PointCLIP \cite{pointclip} further extends the understanding of 3D data by utilizing the ideas of CLIP \cite{clip} to construct an alignment of 2D images with text data. The CLIP \cite{clip} trains the image encoder and the text encoder through contrastive learning, so that the matching image and text pairs are closer in the embedding space, and vice versa. Seal \cite{seal} utilizes VFMs for automotive point cloud segmentation, distilling semantic perceptions from VFMs to point clouds via hyperpixel-driven contrastive learning on camera views. However, fine-grained segmentation requires recognizing tiny structural changes within objects, demanding deep understanding of 3D shapes alongside the ability to capture local details and textual context. The aforementioned methods \cite{pointclip,seal} rely on an indirect alignment strategy using 2D images as intermediaries between 3D point clouds and other data. This approach is highly susceptible to errors in camera intrinsic/extrinsic parameter computation and to mismatches between 2D pixels and 3D points. Consequently, existing methods fail to establish stable, accurate coordinate correspondence between 3D point clouds and 2D images, thus lacking the capability required for point-level fine-grained tasks.

To address these limitations, we propose MR-COSMO, a coarse-to-fine query-driven 3D segmentation model featuring a visual-text memory recall module and direct cross-modal alignment module. Figure \ref{fig1} illustrates the inputs and outputs of the proposed coarse-to-fine model. Our framework first leverages cross-modal mamba-based attention to achieve simultaneous direct alignment between 3D features and both text/2D features, further constrained by contrastive learning applied to the 2D-text pairs. The aligned visual features are then updated via a multilayer multiscale Transformer block and processed through a detection header to generate a 3D bounding box for the input text query. Additionally, we introduce a Memory Module that stores confidence-weighted text-visual feature pairs derived from bounding box contents. This enables effective utilization of prior knowledge during subsequent scene processing, compensating for segmentation inconsistencies across scenarios. Finally, text queries and updated point features within the 3D bounding box are processed through the Memory Module, with segmentation masks generated by an iterative binary classifier. Experiments on diverse indoor and outdoor datasets demonstrate that our method has superior performance over existing methods across multiple downstream tasks. The main contributions of our work can be summarized as follows:

\begin{itemize}
    \item We propose an innovative coarse-to-fine query-driven architecture that first generates coarse 3D object proposals and then performs query-specific fine segmentation, resolving boundary ambiguity in complex scenes.
    % We propose an innovative coarse-to-fine query-driven 3D segmentation network that first coarsely detects 3D objects then performs query-specific fine segmentation.
    % We propose an innovative interactive point cloud segmentation network that first detects 3D objects at coarse scale, then performs query-specific fine segmentation.
    % We propose an innovative interactive point cloud segmentation network that first coarsely detects 3D objects and then finely segments the corresponding query objects.
    \item We introduce Direct Cross-Modal Alignment establishing 3D-text/3D-2D correspondence with Mamba-based attention, enhanced by contrastive learning constraints to overcome camera geometry dependencies.
    % We introduce Direct Cross-Modal Alignment establishing 3D-text/3D-2D correspondence via Mamba-based attention, enhanced with contrastive learning constraints.
    % We introduce a Direct Cross-Modal Alignment module that establishes direct feature correspondence between 3D point clouds and text/2D inputs through the proposed mamba-based attention mechanism, reinforced by contrastive learning constraints.
    % We present a Direct Cross-Modal Alignment module between 3D point clouds and other inputs, directly aligning the 3D features with other data using proposed attention module, with constraints using contrastive learning.
    \item We develop a Memory Module that stores weighted visual-text feature pairs and dynamically recalls prior knowledge via attention mechanisms during inference, ensuring cross-scenario segmentation consistency.
    % We develop a Memory Module storing visual-text feature pairs to leverage retained knowledge for cross-scenario segmentation consistency.
    % We propose a Memory Module that stores multimodal visual-text feature pairs, enabling dynamic utilization of retained knowledge during new scene processing to enhance cross-scenario segmentation consistency.
    % We introduce a Memory Module that stores multimodal visual-text feature pairs, enabling the network to dynamically leverage prior knowledge during new scene processing, thereby improving cross-scenario consistency.
    \item Our method demonstrates state-of-the-art performance across diverse 3D segmentation tasks, confirming robustness under varying textual queries and scene complexities through comprehensive experiments.
    % Our method demonstrates state-of-the-art query-driven segmentation performance across diverse downstream 3D tasks, confirming its robustness.
    % Extensive experiments on diverse 3D instruction, referring, and semantic segmentation datasets demonstrate our method's state-of-the-art query-driven segmentation performance, validating its effectiveness and robustness.
    % Extensive experiments on several indoor and outdoor datasets containing 3D instruction, referring, and semantic segmentation tasks demonstrate that the proposed method has excellent interactive segmentation results, validating the effectiveness and robustness of our work.
\end{itemize}

\section{Related Work}
% \label{sec:related_work}
% In this section, we briefly overview related works of deep learning on point cloud segmentation and text-guided query-driven 3D segmentation.

\subsection{Deep Learning on Point Cloud Segmentation}
% \label{sec:dl-on-pcs}
Recent deep learning advances have significantly progressed point cloud segmentation through novel network architectures and feature learning. By methodology and data format, mainstream approaches include: voxel-based methods, projection-based methods, 2D processing guided methods, MLP-based methods, point convolution-based methods, and attention-based methods.
% \cite{pt,pct,fpt,st,ptv2,superpt,spotr,octformer,sphereformer,supercluster,ptv3,feastmamba}.

Specifically, voxel-based methods \cite{subsc,4dstc} discretize 3D space into regular grids for 3D convolution easily but face resolution-efficiency tradeoffs. Projection-based methods \cite{squeezeseg,polarnet} convert points to 2D representations leveraging established CNNs. 2D processing guided methods \cite{3dmv,waffleiron} adapt 2D networks/VFMs to reduce training costs at the expense of accuracy from domain gaps. MLP-based methods \cite{pointnet,pointnet++} operate directly on raw point clouds, effectively enlarging receptive fields with constrained computation. Point convolution-based methods \cite{pointcnn,kpconv} excel at capturing fine-grained geometry via deformable kernels, yet struggles with efficiency and parameter sensitivity at scale. Recent attention-based advances \cite{pt,pct,st,ptv2,superpt,sphereformer,ptv3,feastmamba} via Transformer and Mamba networks further improve segmentation accuracy.

Based on the current application of deep learning in point cloud segmentation, we still choose the attention network based on Transformer as the backbone.

\begin{figure*}[t]
\centering
% \fbox{\rule{0pt}{2in} \rule{.9\linewidth}{0pt}}
\includegraphics[width=0.9\textwidth]{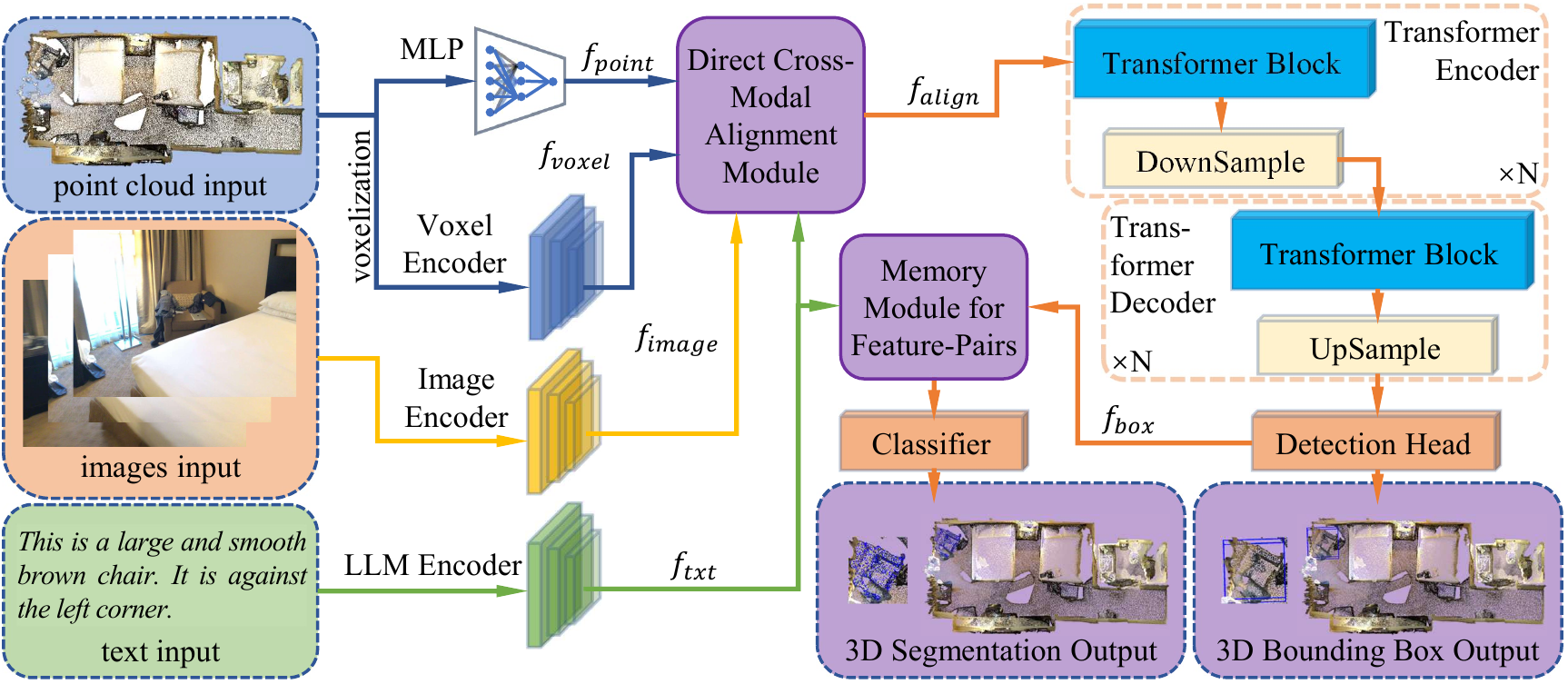}
\caption{A general overview of the proposed network. Given point cloud, images, and text input, the point cloud is first voxelized. Then we use MLP, voxel encoder, image encoder, and LLM encoder to extract four distinct feature representations: $f_{point}$, $f_{voxel}$, $f_{image}$, and $f_{txt}$. These heterogeneous features are then unified via our novel Direct Cross-Modal Alignment (DCMA) module to generate aligned features $f_{align}$. Subsequently, the aligned features undergo refinement through a multi-layer Transformer encoder-decoder architecture. Following this, a detection head produces bounding box predictions and extracts point-wise features $f_{box}$ within each detected region. The $f_{box}$ and $f_{txt}$ are jointly fed into the proposed Memory Module, which leverages stored cross-modal mappings as prior knowledge. Finally, an additional classifier processes the fused features to yield query-driven segmentation results.}
\label{fig2}
\end{figure*}

\subsection{Text-Guided Query-Driven 3D Segmentation}
% \label{sec:tgqd3ds}
Text-guided query-driven 3D segmentation enables natural language-based manipulation and segmentation of 3D objects/scenes, essential for autonomous driving and embodied intelligence requiring precise scene understanding. By leveraging multimodal inputs (text/images/point clouds), existing methods \cite{referit3d,scanrefer,tggnn,butddt,multi3drefer,eda,3dstmn,refmask3d,xrefseg3d} bridge linguistic semantics with 3D geometry, enhancing accuracy in complex environments.
% The goal of the text-guided interactive 3D segmentation task is to enable users to manipulate and segment 3D objects or scenes using natural language commands or prompts. This task is critical for applications such as autonomous driving, robotics, and augmented reality, which require accurate 3D scene understanding and dynamic editing. By utilizing multimodal inputs (e.g., text, images, and point clouds), these methods \cite{referit3d,scanrefer,refmask3d,xrefseg3d,tggnn,butddt,3dstmn,eda,multi3drefer} connect semantic linguistic descriptions with geometric 3D representations, improving usability and accuracy in complex environments.

Specifically, text-guided 3D segmentation encompasses referring and instruction tasks. Referring segmentation outputs masks for objects specified by explicit category words. Instruction segmentation \cite{segpoint} processes functional descriptions without category words. Both require textual comprehension and world knowledge reasoning, yet accuracy lags semantic segmentation, with direct 3D-data alignment still limited.
In this regard, we design a Direct Cross-Modal Alignment module to establish direct correspondence between 3D features and other modalities, addressing errors from parametric computation and inaccurate 2D-3D mappings. Complementarily, we propose a Memory Module that leverages text-visual mapping knowledge to enhance correspondence sensitivity and segmentation accuracy.

% \begin{figure*}
%   \centering
%   \begin{subfigure}{0.32\linewidth}
%     % \fbox{\rule{0pt}{2in} \rule{.9\linewidth}{0pt}}
%     \includegraphics[width=0.95\textwidth]{fig3a}
%     \caption{Framework of the proposed Direct Cross-Modal Alignment module.}
%     \label{fig3-a}
%   \end{subfigure}
%   \hfill
%   \begin{subfigure}{0.64\linewidth}
%     % \fbox{\rule{0pt}{2in} \rule{.9\linewidth}{0pt}}
%     \includegraphics[width=0.95\textwidth]{fig3b}
%     \caption{Framework of the proposed Memory Module for feature pairs.}
%     \label{fig3-b}
%   \end{subfigure}
%   \caption{Visualization of two important modules in the proposed MR-COSMO.}
%   \label{fig3}
% \end{figure*}

\section{Methodology}
% \label{sec:methodology}
% In this section, we first present a general overview of the proposed network framework in Section \ref{sec:framework_overview}. Then we introduce the two important modules of our proposal, Direct Cross-Modal Alignment and Memory Module for Feature Pairs, respectively, in Sections \ref{sec:dcma} and \ref{sec:mmfp} in turn. Finally, we show the objective function of the proposed method in Section \ref{sec:object_function}.

% \subsection{Overview}
% \label{sec:framework_overview}
% The framework of our coarse-to-fine query-driven point cloud segmentation network with a Direct Cross-Modal Alignment module and a Memory Module is shown in Figure \ref{fig2}.
The framework of our MR-COSMO, featuring coarse-to-fine query-driven segmentation with Direct Cross-Modal Alignment and Memory Module, is shown in Figure \ref{fig2}. We process point cloud, corresponding 2D images, and text query as inputs for each individual query-driven segmentation scenario. For point clouds, we employ dual feature extraction: per-point features are extracted via MLP, while voxelized point clouds are processed through a 4-layer 3D transformer with window-shifting to obtain voxel features. For 2D images, visual features are extracted using a pretrained ResNet-50. Text features are generated with LLaMA2-7B \cite{llama2} after vectorizing queries. Finally, our Direct Cross-Modal Alignment module integrates these multimodal features.

After obtaining the aligned features, the features continue to be updated through a multi-layer multi-scale Transformer network. Subsequently, we obtain the bounding box of the target object corresponding to the text query in 3D space through a detection header. The 3D point features within the detected bounding boxes and their associated text query features are jointly input into the proposed Memory Module. Leveraging high-confidence text-visual feature pairs stored in the Memory Module, we iteratively train a binary classifier through an extra loss function. This enables progressive refinement from coarse bounding boxes to precise segmentation masks, establishing our coarse-to-fine framework, as illustrated in Figure \ref{fig1}.

%
% \subsection{Two-Stage Detection-Segmentation network}
% \label{sec:2sdsn}
%

\subsection{Direct Cross-Modal Alignment}
\begin{figure*}[t]
\centering
% \fbox{\rule{0pt}{2in} \rule{.9\linewidth}{0pt}}
\includegraphics[width=0.9\textwidth]{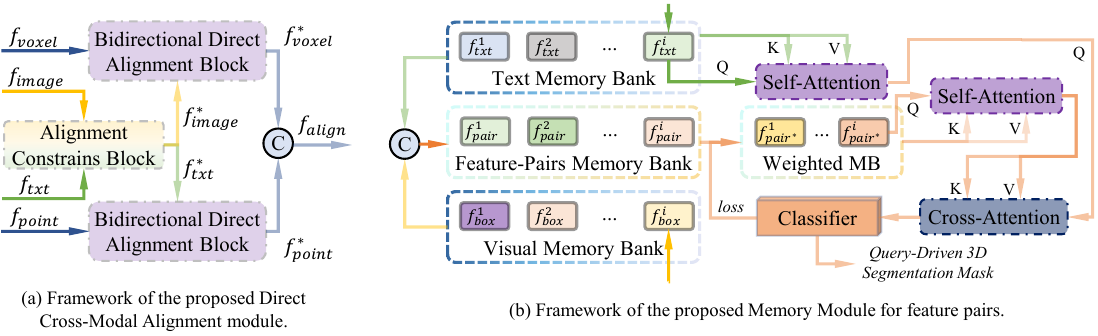}
\caption{Visualization of two important modules in the proposed MR-COSMO.}
\label{fig3}
\end{figure*}

% \label{sec:dcma}
The proposed Direct Cross-Modal Alignment (DCMA) mo-dule, illustrated in Figure \ref{fig3}(a), architecturally consists of two constituent blocks: the Alignment Constrains Block and the Bidirectional Direct Alignment Block. The Alignment Constrains Block enforces contrastive learning constraints between text features and 2D image features, which are directly aligned with 3D features. The Bidirectional Direct Alignment Block implements a novel bidirectional Mamba-based cross-modal attention mechanism to establish direct alignment between 3D features and text/2D image features.

\subsubsection{Alignment Constrains Block}
% \noindent\textbf{Alignment Constrains Block.}
In order to make the subsequent 3D features have the correct correspondence with other modal features in the alignment process, we first process image features $f_{imgae}$ and text features $f_{txt}$ through independent encoders for feature transformation and mapping. This processing enforces matched image-text feature pairs to converge in the embedding space, as shown in Equation \ref{eq1}:
% To ensure precise correspondence between subsequent 3D features and other modalities during alignment, we apply independent encoders to the image features $f_{imgae}$ and text features $f_{txt}$ first for feature processing and mapping, so that the matched image features and text features are closer together in the embedding space, as shown/formalized in Equation \ref{eq1}.

\begin{equation}
  f_{image}^{*},f_{txt}^{*}=\text{Encoders}_{ind}(f_{imgae},f_{txt}).
  \label{eq1}
\end{equation}

For the aforementioned independent image and text encoders, we implement the following training procedure: For each scene in the experimental dataset, we select at most one matching image-text pair, process them through independent encoders to obtain remapped features, and compute cosine similarities between all image-text features to construct a similarity matrix. We then integrate a symmetric cross-entropy loss to maximize similarity scores for correct matching pairs (diagonal elements) while minimizing incorrect pair scores (off-diagonal elements). This contrastive strategy enforces precise alignment between 2D visual and textual representations, mapping corresponding features adjacently in high-dimensional embedding space to constrain subsequent alignment processes.
% We then apply a symmetric cross-entropy loss function to maximize similarity scores for correct matching pairs (diagonal elements) while minimizing scores for incorrect pairs (off-diagonal elements). This contrastive learning strategy enforces alignment between 2D visual and textual representations, effectively mapping corresponding features to adjacent locations in the high-dimensional embedding space to constrain subsequent alignment processes.
% We select at most one pair of matching image and text data for each scene in the experimental dataset, encode them using independent encoders, obtain the remapped image and text features, and compute the cosine similarity between all image and text features to form a similarity matrix. Then a symmetric cross-entropy loss function is introduced to maximize the similarity scores of correct matching pairs (the elements on the diagonal) and minimize the similarity scores of incorrect matching pairs (the off-diagonal elements). This contrastive learning strategy forces the encoder to achieve alignment of 2D and textual representations, effectively mapping the corresponding textual and image features to neighboring locations in the high-dimensional embedding space, so as to achieve the constraint of the following alignment procession.

\subsubsection{Bidirectional Direct Alignment Block}
% \noindent\textbf{Bidirectional Direct Alignment Block.}
After the Alignment Constraints Block, the Bidirectional Direct Alignment Block independently aligns remapped text features $f_{txt}^{*}$ with point features $f_{point}$, and remapped image features $f_{image}^{*}$ with voxel features $f_{voxel}$. This modality pairing offers two key benefits: aligning text directly with points avoids the pixel-to-point misalignment inherent in 2D-3D projection, while aligning image features with voxelized point clouds exploits the regular voxel structure to reduce geometric distortion.
% After the Alignment Constraints Block, we deploy the Bidirectional Direct Alignment Block to independently align remapped text features $f_{txt}^{*}$ with point features $f_{point}$, and remapped image features $f_{image}^{*}$ with voxel features $f_{voxel}$. Our aforementioned modality pairing strategy for feature alignment is motivated by two critical advantages: Aligning text features directly with point features circumvents inherent 2D-3D misalignment errors from pixel-to-point mappings. And pairing image features with voxelized point clouds features leverages the structural regularity of voxel grids to mitigate geometric distortions.

For text-point alignment, given the remapped text features $f_{txt}^{*} \in \mathbb{R}^{1 \times n_d}$ and the neighboring points features $f_{points}^n=[f_{point}^n; f_{point}^{n_1}; \dots; f_{point}^{n_k}] \in \mathbb{R}^{(n_{nbr} + 1) \times n_d}$ for the n-th point with $k$ neighbors, we project both text and neighboring points features into a shared high-dimensional space of dimension $D$:

% \begin{align}
% \phi_{\txt} &= f_{\txt}^{*} W_{\txt} \quad \text{where} \quad W_{\txt} \in \R^{d_{\txt} \times d} \label{eq:proj_txt} \\
% \phi_{\points} &= f_{\points}^n W_{\point} \quad \text{where} \quad W_{\point} \in \R^{d_{\point} \times d} \label{eq:proj_point}
% \label{eq2}
% \end{align}

\begin{equation}
\phi_{txt} = f_{txt}^{*} W_{txt},
\label{eq2}
\end{equation}
%  \quad \text{where} \quad W_{txt} \in \mathbb{R}^{n_d \times D}

\begin{equation}
\phi_{points} = \text{MeanPool}\left(f_{points}^n W_{point}\right),
\label{eq3}
\end{equation}
% \quad \text{where} \quad W_{point} \in \mathbb{R}^{n_d \times D}

\noindent where $W_{txt} \in \mathbb{R}^{n_d \times D}$, $W_{point} \in \mathbb{R}^{n_d \times D}$, $\phi_{txt} \in \mathbb{R}^D$ is the projected text feature, $\phi_{points} \in \mathbb{R}^D$ is the aggregated point feature. And $\text{MeanPool}$ denotes average pooling along the point dimension, we apply this operation to align the projected points features with the projected text features. We then construct a three-element sequence containing the projected text feature $\phi_{txt}$, the aggregated point feature $\phi_{points}$, and a copy of the text feature $\phi_{{txt}^{copy}}$:

% \begin{equation}
% X = \begin{bmatrix}
% \phi_{txt} \\
% \phi_{points} \\
% \phi_{txt}^{copy}
% \end{bmatrix} \in \mathbb{R}^{3 \times D},
% \label{eq4}
% \end{equation}

\begin{equation}
X = \begin{bmatrix}
\phi_{txt} \\
\phi_{points} \\
\phi_{{txt}^{copy}}
\end{bmatrix} \in \mathbb{R}^{3 \times D},
\label{eq4}
\end{equation}

\noindent where sequence $X$ is processed through bidirectional state space models. The forward state space model processes the sequence from top to bottom ($\phi_{txt} \rightarrow \phi_{points} \rightarrow \phi_{{txt}^{copy}}$):

\begin{equation}
h_t^f = \tilde{A}_f h_{t-1}^f + \tilde{B}_f X_t \quad \text{for} \quad t=1,2,3,
\label{eq5}
\end{equation}

% \begin{equation}
% h_{points}^f = \tilde{A}_f h_{txt}^f + \tilde{B}_f \phi_{points},
% \label{eq5}
% \end{equation}

\begin{equation}
\psi_t^f = \tilde{C}_f h_t^f + \tilde{D}_f X_t \quad \text{where} \quad h_0^f = \mathbf{0} \in \mathbb{R}^D,
\label{eq6}
\end{equation}

% \begin{equation}
% \psi_{points}^f = \tilde{C}_f h_{points}^f + \tilde{D}_f \phi_{points},
% \label{eq6}
% \end{equation}
%  \quad \text{where} \quad h_0^f = \mathbf{0} \in \mathbb{R}^D

\noindent The superscripts/subscripts $f$ and $b$ denote forward and backward state-space processing, respectively. The backward model mirrors the forward one but operates on the reversed sequence $(\phi_{txt^{copy}} \rightarrow \phi_{points} \rightarrow \phi_{txt})$. The matrices $\tilde{A}, \tilde{B}, \tilde{C}, \tilde{D} \in \mathbb{R}^{D \times D}$ are learnable parameters. The forward and backward outputs at each position are $\psi_t^f, \psi_t^b \in \mathbb{R}^D$. Specifically, $\psi_1^f$ and $\psi_3^b$ encode pure text features after initial semantic processing; $\psi_2^f$ and $\psi_2^b$ represent text-guided point features; and $\psi_3^f$ and $\psi_1^b$ capture text refined through cross-modal interaction. The final representation is obtained by summing the terminal forward/backward outputs $(\psi_3^f + \psi_1^b)$ followed by Layer Normalization:

% \noindent where the superscript and subscript $f$ stands for the forward process, while the backward state space model processes is similar to the forward one and use superscript and subscript $b$ to separate, but the sequence from bottom to top ($\phi_{{txt}^{copy}} \rightarrow \phi_{points} \rightarrow \phi_{txt}$), and $\tilde{A}, \tilde{B}, \tilde{C}, \tilde{D} \in \mathbb{R}^{D \times D}$ are learnable parameters. $\psi_t^f \in \mathbb{R}^D$ is the output features at each position in the forward processing, while $\psi_t^b \in \mathbb{R}^D$ stands for the backward output features. In particular, the output $\psi_1^f, \psi_3^b$ represents pure textual information after initial semantic encoding, capturing linguistic features before any interaction with point cloud data; $\psi_2^f, \psi_2^b$ embodies text-guided point features where geometric characteristics are modulated by preceding textual context; and $\psi_3^f, \psi_1^b$ reflects text-after-interaction representation containing linguistic features refined through cross-modal processing with point cloud data. Then add the forward and backward outputs $(\psi_3^f + \psi_1^b)$, and apply Layer Normalization to the sum:

\begin{equation}
\psi_{point}^{*} = \text{LayerNorm}(\psi_3^f + \psi_1^b),
\label{eq7}
\end{equation}

\noindent and finally we get the $f_{point}^{*}$ by applying an additional projection layer to project $\psi_{point}^{*}$ back to the dimension $\mathbb{R}^{(n_{nbr} + 1) \times n_d}$ consistent with the input.

Bidirectional direct alignment between voxel features $f_{voxel}$ and remapped image features $f_{image}^{*}$ is similar to the above process, getting voxel features $f_{voxel}^{*}$ that align the image input. Subsequently, the features of the corresponding voxel for each 3D point are concatenated with the point features according to the correspondence between the 3D point and the 3D voxel to obtain the alignment features $f_{align}$ of each 3D point.

\subsection{Memory Module for Feature Pairs}
% \label{sec:mmfp}
Current datasets frequently exhibit imbalanced training sample distributions and inherent intra-class divergence in texture and contextual features across same-category objects. These issues result in the misclassification of objects within the same category and reduced accuracy for categories with a low number of samples. To address these challenges, we propose a Memory Module integrated into the segmentation result generation pipeline, as shown in Figure \ref{fig3}(b).
% Current datasets often suffer from imbalanced training sample distributions compounded by inherent intra-class divergence in texture and contextual features across objects of the same category. These issues lead to misclassifications of same-category objects and reduced accuracy for low-sample categories. To address these challenges, we propose a Memory Module integrated into the segmentation result generation pipeline, as shown in Figure \ref{fig3}(b).

This module stores high-confidence (low-loss) visual-textual feature pairs to constrain the binary classifier generating segmentation results. The module receives text features $f_{txt}^i$ and detection head-processed 3D point features $f_{box}^i$ within bounding boxes as inputs. These features are stored in dedicated text ($\mathcal{M}_t$) and visual ($\mathcal{M}_v$) memory banks alongside prior scene features. We then concatenate corresponding elements to form feature-pairs memory bank:

\begin{equation}
\mathcal{M}_p = \{ [f_{txt}^i; f_{box}^i] \mid i=1,\dots,N \}.
\label{eq8}
\end{equation}

Each pair is assigned an initial weight $w_i^{(0)}=1.0$. To address weighting bias, we introduce both initial and bias weighting. The initial weight is computed from mask loss:

\begin{equation}
w_i^{(\mathrm{init})} = \frac{1}{\mathcal{L}_{BCE_i} + \tau},
\label{eq9}
\end{equation}

\noindent and all samples within the same category $C$ are normalized whenever a new sample is added. This produces the final category-balanced weight:

\begin{equation}
w_i =
\frac{1}{\mathcal{L}_{BCE_i} + \tau}
\cdot
\frac{1}{\sum_{j \in C} \frac{1}{\mathcal{L}_{BCE_j} + \tau}},
\label{eq9a}
\end{equation}

\noindent ensuring each sample’s contribution equals its initial weight divided by the category’s total initial weights. The weighted feature-pairs are then stored as:

% \noindent with each pair assigned an initial weight $w_i^{(0)} = 1.0$ before training begins. These weights are subsequently updated to:

% \begin{equation}
% w_i = \frac{1}{\mathcal{L}_{{BCE}_i} + \tau},
% \label{eq9}
% \end{equation}

% \noindent which is inversely proportional to its object's mask loss $\mathcal{L}_{{BCE}_i}$ (Equation \ref{eq16}), creating weighted feature-pairs:

\begin{equation}
\widetilde{\mathcal{M}}_p = \{ w_i \cdot [f_{txt}^i; f_{box}^i] \}.
\label{eq10}
\end{equation}

During each new scene, the model retrieves stored knowledge through three attention steps: (1) text self-attention takes $f_{txt}^{current}$ as Query and $\mathcal{M}_t$ as Key/Value; (2) feature-pairs self-attention uses $[f_{txt}^{current}; f_{box}^{current}]$ as Query and $\widetilde{\mathcal{M}}_p$ as Key/Value; and (3) cross-attention takes the text self-attention output as Query and the feature-pairs output as Key/Value. The cross-attention result is passed to the binary classifier to generate masks, and its BCE loss updates the current feature-pair’s weight following Equation \ref{eq9a}, enabling dynamic weight refinement that optimizes memory influence and classifier training.

\subsection{Object Function}
% \label{sec:object_function}
The overall network training objective consists of two main components: (1) the combined detection loss $\mathcal{L}{det}$ and segmentation loss $\mathcal{L}{seg}$ for full-network optimization (Equation \ref{eq11}), where $\mathcal{L}{det}$ includes Smooth L1 (Equation \ref{eq14}) and weighted cross-entropy (Equation \ref{eq15}) losses, and $\mathcal{L}{seg}$ uses binary cross-entropy (Equation \ref{eq16}); and (2) the symmetric cross-entropy loss $\mathcal{L}_{DCMA}$ for training the Direct Cross-Modal Alignment module (Equation \ref{eq13}). Together, these form the complete objective (Equation \ref{eq12}):
% The overall network training objective function comprises two primary components: (1) the combined detection loss $\mathcal{L}_{det}$ and segmentation loss $\mathcal{L}_{seg}$ for full-network optimization, formalized in Equation \ref{eq11}, where $\mathcal{L}_{det}$ integrates Smooth L1 Loss (Equation \ref{eq14}) and weighted cross-entropy loss (Equation \ref{eq15}), while $\mathcal{L}_{seg}$ employs binary cross-entropy loss (Equation \ref{eq16}); and (2) the specialized symmetric cross-entropy loss $\mathcal{L}_{DCMA}$ exclusively for Direct Cross-Modal Alignment module training, defined in Equation \ref{eq13}. These components collectively constitute the complete objective function as shown in Equation \ref{eq12}:

% The objective function of the overall network training process is mainly divided into the following parts: one part is the detection loss and segmentation loss used for the whole network training process as shown in Equation \ref{eq11}, the detection loss consists of Smooth L1 Loss and weighted cross-entropy loss, and the segmentation loss consists of a binary cross-entropy loss. The other part is a symmetric cross-entropy loss function used only for the training process of the Direct Cross-Modal Alignment module. The above two parts of the objective function constitute that of the whole network, as shown in Equation \ref{eq12}.
% Detailed formulas for each part of the objective function are shown below.

\begin{align}
\mathcal{L}_{task}&=\alpha \mathcal{L}_{det} + \beta \mathcal{L}_{seg}  \notag \\
&=\alpha \mathcal{L}_{smoothL1} + \alpha \mathcal{L}_{WCE} + \beta \mathcal{L}_{BCE},
\label{eq11}
\end{align}

\begin{equation}
\mathcal{L}_{all} = \mathcal{L}_{task} + \mathcal{L}_{DCMA},
\label{eq12}
\end{equation}

% As mentioned above, the objective function of our network consists of the detection loss $\mathcal{L}_{det}$ and segmentation loss $\mathcal{L}_{seg}$ for training the main network as shown in Equation \ref{eq11} and the loss $\mathcal{L}_{DCMA}$ for training the Direct Cross-Modal Alignment module as shown in Equation \ref{eq13}. We apply Smooth L1 loss as well as weighted cross-entropy loss to construct the detection loss which are shown in Equations \ref{eq14} and \ref{eq15}. For segmentation loss, we use a binary cross-entropy loss function as shown in Equation \ref{eq16}:

% \begin{align}
% \mathcal{L}_{task}&=\alpha \mathcal{L}_{det} + \beta \mathcal{L}_{seg}  \notag \\
% &=\alpha \mathcal{L}_{smoothL1} + \alpha \mathcal{L}_{WCE} + \beta \mathcal{L}_{BCE},
% \label{eq9}
% \end{align}

\begin{align}
& \mathcal{L}_{DCMA}=\mathcal{L}_{SCE}  \notag \\
&=\gamma \cdot \left( -\sum_{i=1}^N y_i \log(p_i^{sce}) \right) + \notag \\
&\delta \cdot \left( -\sum_{i=1}^N p_i^{sce} \log(y_i) \right),
\label{eq13}
\end{align}

\begin{equation}
\mathcal{L}_{smoothL1}(y, \hat{y}) = \begin{cases}
0.5(\hat{y} - y)^2/\epsilon, & \text{if } |\hat{y} - y| < \epsilon \\
|\hat{y} - y| - 0.5\epsilon, & \text{otherwise}
\end{cases},
\label{eq14}
\end{equation}

\begin{equation}
\mathcal{L}_{WCE} = -\frac{1}{N}\sum_{i=1}^N \sum_{c=1}^C w_c \cdot y_{i,c} \log(p_{i,c}^{wce}),
\label{eq15}
\end{equation}

\begin{align}
& \mathcal{L}_{BCE} \notag \\
& = -\frac{1}{N}\sum_{i=1}^N \left[ y_i \log(p_i^{bce}) + (1-y_i)\log(1-p_i^{bce}) \right],
\label{eq16}
\end{align}

\noindent where $y_i$ is the true label corresponding to the $i$th sample (where $y_i = 1$ indicates the sample belongs to a specific class, and $0$ otherwise), $p_i^{sce}$ is the predicted probability output by the model for the $i$th sample, $y$ is the true value, $\hat{y}$ is the predicted value, $\epsilon$ is a smoothing threshold parameter, $N$ is the number of samples, $C$ is the total number of categories, $y_{i,c}$ is the true one-hot label of the $i$th sample, $p_{i,c}^{wce}$ is the predicted probability of the $i$th sample for category $c$, $w_c$ is the weight of category $c$, and $p_i^{bce}$ is the probability that the $i$th sample is predicted to be the positive class.

\section{Experiments}
% \label{sec:experiments}

% In this section, we first provide an overview of the experimental settings in Section \ref{sec:settings}. We then present the 3D instruction and referring segmentation results in Sections \ref{sec:instruction-seg} and \ref{sec:referring-seg}, respectively. After that, we conduct 3D semantic segmentation experiments in Section \ref{sec:semantic-seg}. Furthermore, we showcase the extensive ablation study in Section \ref{sec:ablation}.

\subsection{Experimental Settings}
% \label{sec:settings}

\subsubsection{Datasets}
% \label{sec:datasets}
% \noindent\textbf{Datasets.}
% For a variety of 3D segmentation tasks, we use a diverse set of newly ÷、proposed or commonly used publicly available datasets for evaluation.
The proposed network is applicable to diverse point cloud segmentation downstream tasks, with experimental validation conducted on three key tasks: 3D instruction segmentation using Instruct3D \cite{segpoint} built upon ScanNet++ \cite{scannet++}, 3D referring segmentation using ScanRefer \cite{scanrefer} based on ScanNet \cite{scannet}, and 3D semantic segmentation evaluated on both the indoor S3DIS \cite{s3dis} and outdoor SemanticKITTI \cite{semantickitti} datasets. 
% Comprehensive dataset specifications are provided in the Supplementary Material.
% The network we proposed can be applied to a variety of downstream tasks of point cloud segmentation. In the experiment, we selected 3D instruction, 3D referring and 3D semantic segmentation tasks to test the network. For the 3D instruction segmentation, we use Instruct3D \cite{segpoint} based on ScanNet++ \cite{scannet++}; for the 3D referring segmentation task, we use ScanRefer \cite{scanrefer} based on ScanNet \cite{scannet}; in addition, for the semantic segmentation task, we chose a commonly used indoor dataset, S3DIS \cite{s3dis}, and an outdoor dataset, SemanticKITTI \cite{semantickitti}. Additional information about the datasets aforementioned can be found in the Supplementary Material.

\paragraph{3D Instruction Segmentation Datasets.} The Instruct3D dataset \cite{segpoint} constructs a comprehensive evaluation framework using 280 reconstructed indoor scenes sourced from ScanNet++ \cite{scannet++}. This dataset provides 2,565 functional language ins-truction-point cloud pairs, with each scene containing an average of 9.16 linguistically diverse instructions. These instructions specifically require the segmentation network to demonstrate comprehension capabilities that extend beyond basic categorical identification, encompassing advanced understanding of object functionality. The dataset is formally partitioned into 2,052 training pairs and 513 validation pairs to facilitate systematic evaluation. Furthermore, the benchmark introduces significant combinatorial reasoning challenges by requiring the network to simultaneously process three critical dimensions: spatial relationships ("near the window"), functional attributes ("adjustable"), and contextual constraints ("in the office area").

\paragraph{3D Referring Segmentation Datasets.} The ScanRefer dataset \cite{scanrefer} contains 51,583 densely annotated descriptions distributed across 800 ScanNet \cite{scannet} scenes. This collection achieves an average density of 64.5 descriptions per environment. Each annotation incorporates explicit category specifications such as "the red sofa", spatial relations including "left of the doorway", and attribute descriptors like "wooden rectangular table". ScanRefer leverages ScanNet's scene-object ID mapping with 1,213 object categories to establish precise text-to-geometry correspondence. The dataset adopts ScanNet's standardized partitioning scheme, consisting of 1,201 training scenes containing 36,665 references, 312 validation scenes with 9,508 references, and 100 test scenes comprising 5,410 references.

\paragraph{3D Semantic Segmentation Datasets.} The S3DIS dataset \cite{s3dis} comprises 272 complete building scans, encompassing six distinct architectural areas, including offices, lobbies and auditoriums. This comprehensive dataset also includes 11,368 rooms, each meticulously annotated with detailed descriptors. The system provides hierarchical annotations for 13 structural categories, which are organized into four groups: structural elements (ceiling, floor, wall), architectural components (beam, column, window, door), movable objects (table, chair, sofa, bookcase, board), and miscellaneous clutter. The SemanticKITTI dataset \cite{semantickitti} represents an extension of the KITTI Vision Benchmark \cite{kitti} through the augmentation of raw LiDAR sequences with per-point semantic annotations. This augmentation provides sequential point clouds captured during 22 urban drives, with a total distance covered of 43 kilometres. The benchmark under scrutiny in this study exhibits three critical characteristics. Firstly, it demonstrates temporal coherence with over 4,500 sequential frames. Secondly, it contains dynamic scenes containing moving objects. Thirdly, it exhibits a long-tail distribution across 20 semantic classes, including road and motorcycle. Sequence partitioning allocates 19,130 training frames from sequences 00-07 and 09-10, 4,071 validation frames from sequence 08, and 20,351 test frames from sequences 11-21.

Query Generation for Semantic Segmentation In order to evaluate our query-driven 3D segmentation method on semantic segmentation using the S3DIS and SemanticKITTI datasets, a standardized text query generation methodology has been implemented. The approach described here employs the following structured template: \textit{‘Please segment the masks of all the categories in this point cloud in order and output separate masks for each category in the following order: \{category1\}, \{category2\}, ...’}. For S3DIS, the category sequence is delineated as follows: \textit{ceiling, floor, wall, beam, column, window, door, table, chair, sofa, bookcase, board, clutter}. This sequence necessitates that the model parses 13 indoor categories, generates instance-level masks in textual order. For SemanticKITTI, the sequence is delineated as follows: \textit{car, bicycle, motorcycle, truck, other-vehicle, person, bicyclist, motorcyclist, road, parking, sidewalk, other-ground, building, fence, vegetation, trunk, terrain, pole, traffic-sign}. This sequence demands the processing of 19 outdoor classes. This methodology enables precise semantic segmentation in complex 3D environments through structured commands. These commands activate the network's cross-modal alignment capabilities during segmentation processing, maintaining geometric-semantic consistency. The methodology integrates with our memory module to leverage retrieved visual-text knowledge, resolving categorical ambiguities.

\subsubsection{Metrics}
% \label{sec:metrics}
% \noindent\textbf{Metrics.}
Following established practices in 3D segmentation research, we adopt standard mean Intersection-over-Union (mIoU) for main evaluation. For the Instruct3D ben-chmark \cite{segpoint}, we employ SegPoint's accuracy (Acc) metric, which defines a sample as correctly identified if IoU $>$ 0.5 and calculates accuracy as the proportion of correctly identified samples relative to all samples. The best indicators are highlighted in \textbf{bold} and the next best indicators are \underline{underlined}.

%
% % \subsubsection{Comparison Methods}
% % \label{sec:methods}
% \noindent\textbf{Comparison Methods.}
%

\begin{figure*}[t]
\centering
% \fbox{\rule{0pt}{2in} \rule{.9\linewidth}{0pt}}
\includegraphics[width=0.95\textwidth]{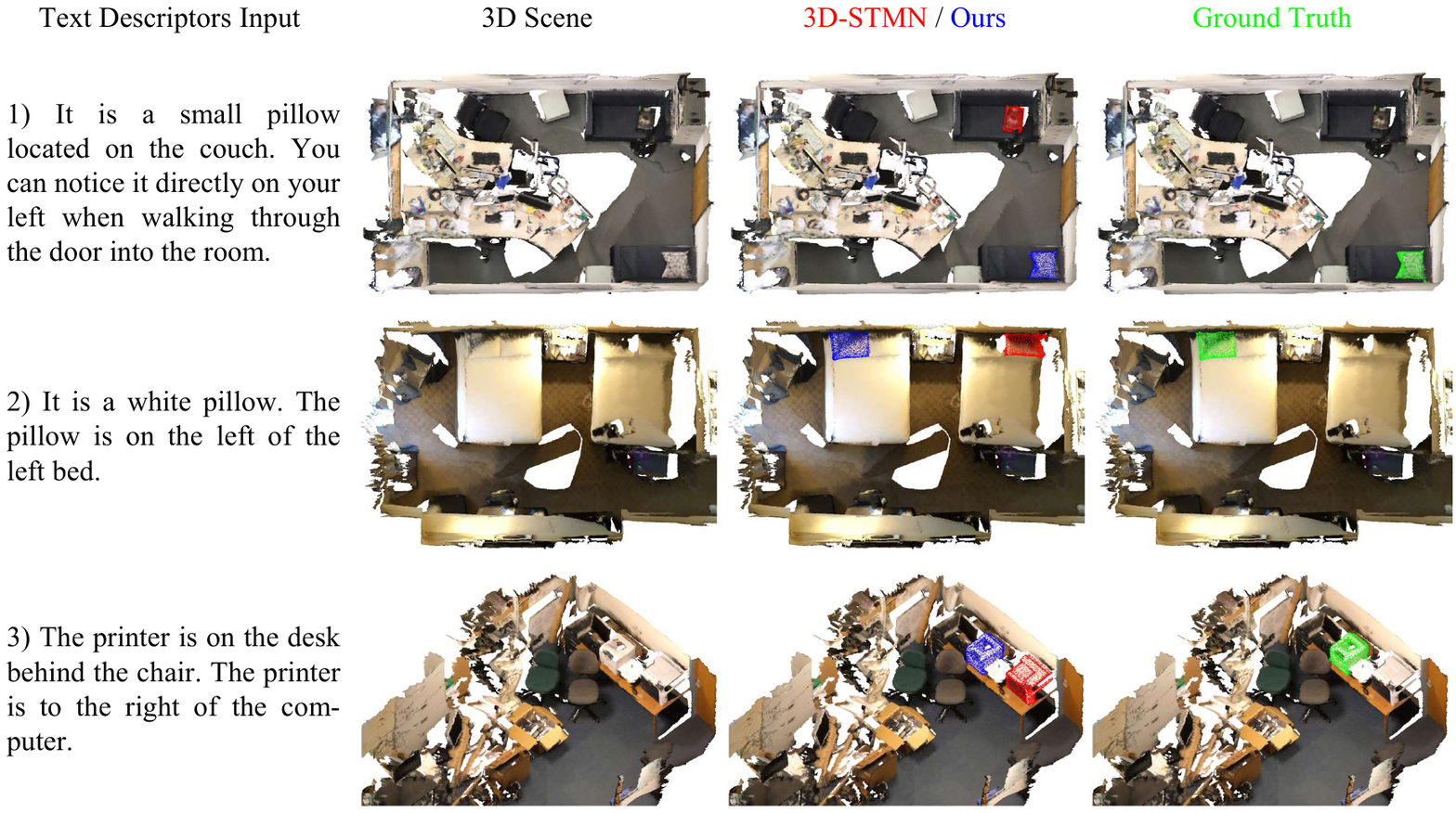}
\caption{Qualitative results of 3D referring segmentation experiment on ScanRefer \cite{scanrefer}. In the visualization results, the blue masks denote the segmentation outputs from the proposed MR-COSMO, the red masks represent the predictions of the 3D-STMN \cite{3dstmn}, and the green masks indicate the ground truth annotations.}
\label{fig4}
\end{figure*}

\subsubsection{Implementation Details}
% \label{sec:details}
% \noindent\textbf{Implementation Details.}
Experiments are implemented on a server equipped with four Nvidia V100 (32G $\times$ 4 GPU memory). Consistent with PTv3 \cite{ptv3}, we use the AdamW optimizer with a cosine scheduler during training and set 1$\%$ of the training process as a warm-up. We set the initial learning rate and weight decay to 0.005, 0.05 and 0.002, 0.005 for indoor and outdoor scenes, respectively. In addition, the number of training epochs was set to 500 and 100 for the indoor and outdoor scenes, accordingly. For the LLaMa 2-7B model, we employed a pretrained frozen model, feeding only vectorized text inputs to generate standardized textual features. For the memory module storing thousands of text-visual feature pairs from training samples, we converted the feature data type from float32 to float16. This significantly reduced memory consumption while effectively preserving data precision, limiting this component's memory usage to under 50MB. Moreover, the memory module dynamically updates the weights for each training sample. Since the textual prompts encompass not only the target object category information but also details such as the object's colour and texture, its surrounding environment, or its positional relationship with adjacent objects, there is no risk of overfitting. Random seeds employed included the conventional 42, alongside 888 and 2026. Each experiment was run at least three times, with the average result presented in the tables. The standard deviation across multiple runs was consistently controlled below 0.2$\%$.

\begin{table}
  \centering
  {\footnotesize
  \begin{tabular}{l|cc}
    \toprule
    % \makebox[25mm][l]{Method} & \makebox[15mm]{Acc} & \makebox[15mm]{mIoU}\\
    Method & Acc & mIoU\\
    \midrule
    ReferIt3D \cite{referit3d} & 11.7 & 6.4\\
    ScanRefer \cite{scanrefer} & 12.0 & 6.9\\
    TGNN \cite{tggnn} & 12.9 & 7.1\\
    BUTD-DETR \cite{butddt} & 16.3 & 10.9\\
    EDA \cite{eda} & 16.6 & 12.1\\
    M3DRef \cite{multi3drefer} & 18.1 & 12.8\\
    SegPoint \cite{segpoint} & 31.6 & \underline{27.5}\\
    \midrule
    % Ours* & \underline{32.3} & 27.2\\
    Ours* & \underline{31.9} & 27.4\\
    % Ours & \textbf{34.1} & \textbf{28.4}\\
    Ours & \textbf{33.8} & \textbf{28.5}\\
    \bottomrule
  \end{tabular}
  }
  \caption{3D instruction segmentation results on Instruct3D \cite{segpoint}. * denotes our method removing the proposed Memory Module.}
  \label{tab1}
\end{table}

\begin{table}
  \centering
  {\footnotesize
  \begin{tabular}{l|c}
    \toprule
    % \makebox[25mm][l]{Method} & \makebox[18mm]{mIoU}\\
    Method & mIoU\\
    \midrule
    TGNN \cite{tggnn} & 27.8\\
    BUTD-DETR \cite{butddt} & 35.4\\
    EDA \cite{eda} & 36.2\\
    M3DRef \cite{multi3drefer} & 35.7\\
    X-RefSeg3D \cite{xrefseg3d} & 29.9\\
    3D-STMN \cite{3dstmn} & 39.5\\
    RefMask3D \cite{refmask3d} & \underline{44.8}\\
    SegPoint \cite{segpoint} & 41.7\\
    \midrule
    % Ours & \textbf{45.3}\\
    Ours & \textbf{45.6}\\
    \bottomrule
  \end{tabular}
  }
  \caption{3D referring segmentation results on ScanRefer \cite{scanrefer}.}
  \label{tab2}
\end{table}

\subsection{Experiments on 3D Instruction Segmentation}
% \label{sec:instruction-seg}
We present the performance on 3D instruction segmentation task of MR-COSMO and SOTA methods in Table \ref{tab1}. In the Instruct3D \cite{segpoint} experiment, MR-COSMO achieve excellent results, outperforming the SOTA method in both Acc and mIoU evaluation metrics.

Even when not incorporating the proposed Memory Module for storing high-confidence text-visual query pairs, our method has a competitive level of accuracy compared to the best existing method, SegPoint \cite{segpoint}. Compare the last three rows of data in the table, our method without the Memory Module is 0.3$\%$ ahead of SegPoint \cite{segpoint} in Acc while only 0.1$\%$ behind in mIoU, and our full method is 2.2$\%$ and 1.0$\%$ ahead of SegPoint \cite{segpoint} in Acc and mIoU. These results show that our method has a strong text comprehension and reasoning ability.

\subsection{Experiments on 3D Referring Segmentation}
% \label{sec:referring-seg}
% We showcase the performance on 3D referring segmentation task of MR-COSMO and SOTA methods in Table \ref{tab2}. In the ScanRefer \cite{scanrefer} experiment, our MR-COSMO achieves superior results, with at least 0.5$\%$ mIoU improvement over SOTA methods.

We demonstrate MR-COSMO's performance on 3D referring segmentation against SOTA methods in Table \ref{tab2}, achieving superior results on ScanRefer \cite{scanrefer} with at least 0.8$\%$ mIoU improvement. This demonstrates that our method not only has strong comprehension and inference ability, but also has text-visual association construction ability, which means that it can query the distribution of the corresponding objects in 3D space based on the explicit text words.

Moreover, we also conduct qualitative results of our method on the validation set of ScanRefer \cite{scanrefer}, as shown in Figure \ref{fig4}. We select 3D-STMN \cite{3dstmn} as the comparison baseline because RefMask3D \cite{refmask3d} lacks pretrained models and SegPoint \cite{segpoint} has unreleased source code, making 3D-STMN the strongest fully available method for qualitative evaluation. It can be seen that our method has stronger individual discrimination and recognition ability when there are multiple similar objects of the target category in the same space. Unlike 3D-STMN's susceptibility to selecting non-target objects with complex text descriptions, our method maintains accurate target identification through the Memory Module's text-visual mapping knowledge.
% For scenes with multiple objects of the same category in close distance and when the text descriptions are long and complex, 3D-STMN \cite{3dstmn} has a higher probability of taking non-target objects as outputs yielding incorrect target responses. In contrast, our MR-COSMO can better understand and reason to get the correct target object of the text description. This is also attributed to the rich knowledge of text-visual query mapping that the proposed Memory Module brings to the overall network.

\begin{figure*}[!ht]
\centering
% \fbox{\rule{0pt}{2in} \rule{.9\linewidth}{0pt}}
\includegraphics[width=0.95\textwidth]{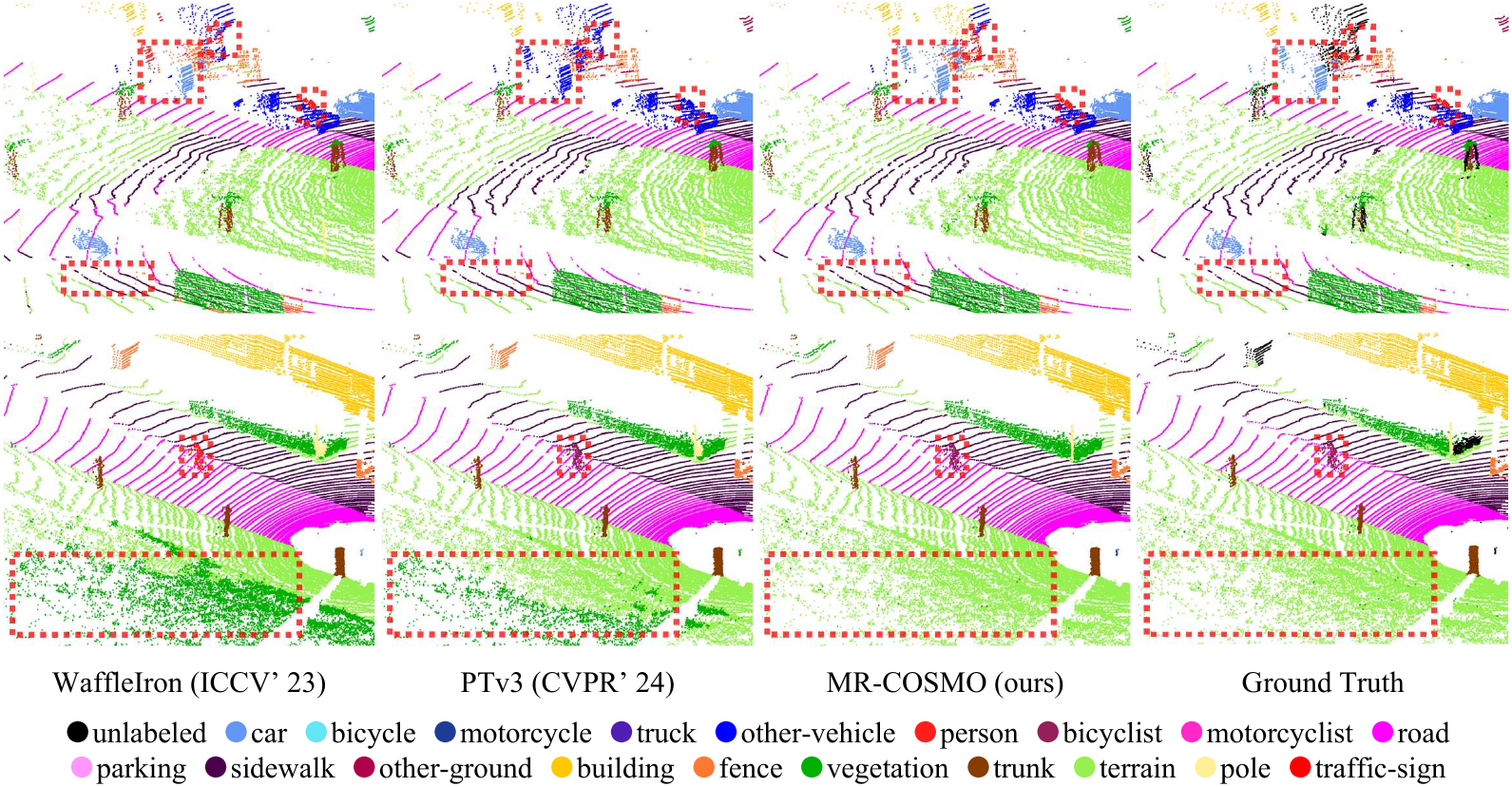}
\caption{Qualitative results of 3D semantic segmentation experiment on SemanticKITTI \cite{semantickitti}.}
\label{fig5}
\end{figure*}

\subsection{Experiments on 3D Semantic Segmentation}
% \label{sec:semantic-seg}
We summarize the performance on 3D semantic segmentation task of MR-COSMO and SOTA methods in Table \ref{tab3} and Table \ref{tab4}. Experiments on the S3DIS \cite{s3dis} Area 5 (Table \ref{tab3}) show that our method outperforms the method \cite{ptv3,ppt} that using multiple datasets as training data in terms of all three overall accuracy metrics. Experiments on the SemanticKITTI validation set (Table \ref{tab4}) show that MR-COSMO has an accuracy improvement of at least 1.1$\%$ in mIoU compared to existing methods. This can prove that allowing the 3D segmentation network to perceive the category information in the scene in advance during the feature processing stage helps its segmentation accuracy to be improved. For a fair comparison, our method is trained using only the data in the corresponding dataset in our experiments, and no data from additional datasets are added to augment the training samples.

\begin{table}
  \centering
  {\scriptsize
  \begin{tabular}{l|ccc}
    \toprule
    % \makebox[28mm][l]{Method} & \makebox[10mm]{OA} & \makebox[10mm]{mAcc} & \makebox[10mm]{mIoU}\\
    Method & OA & mAcc & mIoU\\
    \midrule
    PointNet \cite{pointnet} & - & 49.0 & 41.1\\
    KPConv \cite{kpconv} & - & 72.8 & 67.1\\
    PCT \cite{pct} & - & 67.7 & 61.3\\
    ST \cite{st} & 91.5 & 78.1 & 71.0\\
    SPT \cite{superpt} & 89.5 & 77.3 & 68.9\\
    PTv2 \cite{ptv2} & 91.6 & 78.0 & 72.6\\
    PTv3 \cite{ptv3} & - & - & 73.4\\
    PTv3+PPT \cite{ptv3,ppt} & \underline{92.0} & \underline{80.1} & \underline{74.7}\\
    \midrule
    % Ours & \textbf{93.2} & \textbf{84.5} & \textbf{75.8}\\
    Ours & \textbf{92.8} & \textbf{84.3} & \textbf{75.6}\\
    \bottomrule
  \end{tabular}
  }
  \caption{Indoor 3D semantic segmentation results on S3DIS \cite{s3dis}.}
  \label{tab3}
\end{table}

\begin{table}
  \centering
  {\footnotesize
  \begin{tabular}{l|c}
    \toprule
    % \makebox[25mm][l]{Method} & \makebox[18mm]{mIoU}\\
    Method & mIoU\\
    \midrule
    SphereFormer \cite{sphereformer} & 67.8\\
    WaffleIron \cite{waffleiron} & 68.0\\
    2DPASS \cite{2dpass} & 69.3\\
    PTv2 \cite{ptv2} & 70.3\\
    OA-CNNs \cite{oacnns} & 70.6\\
    PPT+SparseUNet \cite{ppt} & 71.4\\
    PTv3+PPT \cite{ptv3,ppt} & \underline{72.3}\\
    \midrule
    % Ours & \textbf{73.5}\\
    Ours & \textbf{73.4}\\
    \bottomrule
  \end{tabular}
  }
  \caption{Outdoor 3D semantic segmentation results on the validation set of SemanticKITTI \cite{semantickitti}.}
  \label{tab4}
\end{table}

In addition to quantitative comparisons in the main manuscript, we conduct qualitative evaluations on the SemanticKITTI validation set \cite{semantickitti}. As shown in Figure \ref{fig5}, our method demonstrates superior segmentation results in critical regions indicated by red boxes. This qualitative enhancement is particularly evident in the differentiation of perceptually similar categories, such as diverse ground types (terrain vs. vegetation) and structurally analogous vehicles (cars vs. trucks). This capability is derived from the method's scene-specific category perception, which exceeds the performance of conventional 3D semantic segmentation approaches. Furthermore, the Memory Module's learned mappings between category semantics and discriminative visual features enable the network to distinguish between similar categories of objects with greater accuracy.

\subsection{Ablation Experiments}
% \label{sec:ablation}
Compare the last two rows of data shown in Table \ref{tab1}, the method with the proposed Memory Module has an accuracy improvement of 1.9$\%$ and 1.1$\%$ in Acc and mIoU, respectively, which shows the importance of the proposed Memory Module in improving the network's ability to understand text-visual feature correspondences.

\begin{table}
  \centering
  {\scriptsize
  \begin{tabular}{l|c|c}
    \toprule
    % \makebox[15mm][l]{Method} & \makebox[5mm]{mIoU} & \makebox[8mm]{+$\Delta$/-$\Delta$}\\
    Method & mIoU & $\Delta$\\
    \midrule
    % \hline
    % Baseline & 26.4 & +0.0/-2.0\\
    Baseline & 26.4 & +0.0/-2.1\\
    % w/ Direct Cross-Modal Alignment (DCMA) & 27.2 & +0.8/-1.2\\
    w/ Direct Cross-Modal Alignment (DCMA) & 27.4 & +1.0/-1.1\\
    % w/ Memory Module for Feature Pairs (MMFP) & 27.5 & +1.1/-0.9\\
    w/ Memory Module for Feature Pairs (MMFP) & 27.5 & +1.1/-1.0\\
    % \midrule
    \hline
    % Only Point as Q and Text as K/V & 27.0 & +0.6/-1.4\\
    % Only Use Voxel Encoder for Point Cloud & 27.9 & +1.5/-0.5\\
    Only Use Voxel Encoder for Point Cloud & 27.7 & +1.3/-0.8\\
    Reversed matching (text-voxel and image-point) & 27.9 & +1.5/-0.6\\
    % w/o Alignment Constrains Block & 28.1 & +1.7/-0.3\\
    w/o Alignment Constrains Block & 28.0 & +1.6/-0.5\\
    % w/o Loss on Bounding Boxes & 28.2 & +1.8/-0.2\\
    w/o Loss on Bounding Boxes & 28.4 & +2.0/-0.1\\
    % \midrule
    \hline
    % w/ all & 28.4 & +2.0/-0.0\\
    w/ all & 28.5 & +2.1/-0.0\\
    \bottomrule
  \end{tabular}
  }
  \caption{Ablation study of different proposed modules on Instruct3D \cite{segpoint}.}
  \label{tab5}
\end{table}

\begin{table}[t]
\centering
% \resizebox{0.47\textwidth}{!}{
% \renewcommand\arraystretch{0.77}
{\scriptsize
\begin{tabular}{l|cccc}
\toprule
  % \makebox[7mm][l]{Method} & \makebox[5mm]{Speed(fps)} & \makebox[5mm]{GPU(GB)} & \makebox[9mm]{runtime(ms)} & \makebox[3mm]{mIoU}\\\hline
  Method & Speed(fps) & GPU(GB) & runtime(ms) & mIoU\\\hline
  3D-STMN & 3.53 & 31 & 283 & 39.5\\
  Ours & 2.66 & 28.9 & 376 & 45.6\\
\bottomrule
\end{tabular}
% }
}
\caption{Runtime evaluation on ScanRefer \cite{scanrefer}.}
\label{tab6}
\end{table}

%  backbone choices, and hyperparameter settings

\begin{table}[t]
\centering
% \resizebox{0.47\textwidth}{!}{
% \renewcommand\arraystretch{0.77}
{\scriptsize
\begin{tabular}{l|ccc}
\toprule
  Method & Speed(fps) & GPU(GB) & mIoU\\\hline
  Replace Mamba with Transformer & 2.21 & 30.4 & 28.4\\
  Replace ResNet50 with ViT & 1.89 & 33.7 & 28.7\\
  Replace LLaMA2-7B with 13B & 2.35 & 33.0 & 28.6\\
  Replace LLaMA2-7B with 2B & 2.74 & 27.6 & 28.2\\
  \hline
  Ours & 2.66 & 28.9 & 28.5\\
\bottomrule
\end{tabular}
% }
}
\caption{Runtime and mIoU for Different Backbone Architectures on Instruct3D \cite{segpoint}}
\label{tab7}
\end{table}

\begin{table}[t]
\centering
% \resizebox{0.47\textwidth}{!}{
% \renewcommand\arraystretch{0.77}
{\scriptsize
\begin{tabular}{l|c}
\toprule
  The value of $\tau$ in Equations \ref{eq9} and \ref{eq9a} & mIoU\\\hline
  $\tau = 0.01$ & 27.6\\
  $\tau = 0.1$ & 28.1\\
  \hline
  $\tau = 0.05$ & 28.5\\
\bottomrule
\end{tabular}
% }
}
\caption{Ablation study of the hyperparameter $\tau$ on Instruct3D \cite{segpoint}}
\label{tab8}
\end{table}

Table \ref{tab5} summarizes the impact of the proposed Direct Cross-Modal Alignment and Memory Module for Feature Pairs on Instruct3D \cite{segpoint}. The experimental results demonstrate that network accuracy achieves enhancements of 1.0$\%$, 1.1$\%$, and 2.1$\%$ under three conditions: (a) introducing the Direct Cross-Modal Alignment module, (b) deploying the Memory Module for Feature Pairs independently, and (c) combining both modules, respectively.The proposed method mitigates the negative impacts caused by error accumulation and 2D-3D coordinate mapping distortions in conventional alignment approaches through two strategic mechanisms: (1) explicitly bridging 3D features with multimodal representations via direct alignment, and (2) integrating text-2D feature correlations by contrastive learning as constraints to regulate the 3D feature alignment process. Meanwhile, the proposed Memory Module can further constrain the output process from 3D bounding boxes to segmentation results, and store high-confidence text-to-vision mapping relations to enhance the text-to-vision inference capability of the network. Combining both modules further improves accuracy in interactive segmentation tasks. In addition, we show that using both voxels and point MLP features for point cloud processing provides a 0.8$\%$ accuracy advantage over using voxels alone. We also demonstrate that applying contrastive learning for modal alignment improves the network’s accuracy by 0.5$\%$, highlighting the contribution of each module to overall performance. Finally, ablating the constraints on the detection frame results in only a 0.1$\%$ change in accuracy, which helps the network discriminate between individuals of the same class without unfairly affecting comparisons with other methods.

Also, Table \ref{tab5} shows that using only voxel features for point cloud processing achieves 27.7$\%$ mIoU, while reversed matching between text-voxel and image-point features improves it slightly to 27.9$\%$. Removing the Alignment Constraints Block results in 28.0$\%$ mIoU, and excluding the bounding box loss further increases performance to 28.4$\%$, suggesting that each design choice contributes to improved multimodal alignment.  

Table \ref{tab6} reports the runtime evaluation on ScanRefer \cite{scanrefer}. Our method achieves 45.6$\%$ mIoU with a speed of 2.66 fps and 28.9 GB GPU memory, while 3D-STMN reaches 39.5$\%$ mIoU at 3.53 fps using 31 GB GPU memory. This indicates that our method improves accuracy with a moderate trade-off in speed and memory consumption.  

Table \ref{tab7} presents results for different backbone architectures on Instruct3D. Replacing Mamba attention with a standard Transformer reduces mIoU to 28.4, while using ViT instead of ResNet50 slightly increases it to 28.7$\%$. Scaling LLaMA2 from 7B to 13B yields 28.6$\%$ mIoU, whereas the smaller 2B version drops to 28.2$\%$. Our default configuration (Mamba + ResNet50 + LLaMA2-7B) achieves 28.5$\%$ mIoU with a speed of 2.66 fps and 28.9 GB GPU usage, confirming that the chosen backbones balance performance and efficiency.  

Finally, Table \ref{tab8} investigates the impact of the temperature hyperparameter $\tau$ in Equations \ref{eq9} and \ref{eq9a}. Setting $\tau = 0.05$ achieves the best mIoU of 28.5$\%$, while values of 0.01 and 0.1 yield 27.6$\%$ and 28.1$\%$, respectively, highlighting the sensitivity of contrastive learning to temperature settings.  

Overall, the experimental results demonstrate that each proposed module, backbone choice, and hyperparameter selection contributes to improved 3D interactive segmentation performance, with our method achieving a strong balance of accuracy, runtime, and memory efficiency.

\section{Conclusions}
% \label{sec:conclusions}
We propose MR-COSMO, a coarse-to-fine query-driven 3D segmentation model integrating Memory Recall and Direct Cross-Modal Alignment. This framework mitigates computational errors from intrinsic/extrinsic parameters through a Direct Cross-Modal Alignment module establishing explicit 3D feature alignment with text/image modalities. Complementarily, a Memory Module stores high-confidence text-visual feature mappings during training to leverage prior knowledge for enhancing segmentation accuracy when processing new scenes. Comprehensive experiments across diverse segmentation tasks demonstrate significant performance gains over comparative methods with consistent cross-scene generalization, confirming the method's effectiveness and robustness.

\section{Acknowledgments}
This work is supported by the National Key Research and Development Program of China under Grant No. 2024YFB4709100, the National Natural Science Foundation of China under Grant No. 62572468 and Beijing Natural Science Foundation under Grant No. L241012. We thank the anonymous Program Committees and Program Chairs so much for their helpful comments and suggestions.

\bibliography{aaai2026}

\end{document}